\documentclass{article}
\usepackage{spconf,amsmath,amssymb,graphicx,booktabs,multirow,tikz}
\usepackage{float}
\usepackage{stfloats}
\usepackage{url}
\usetikzlibrary{arrows.meta,positioning,fit,backgrounds,calc}

\newcommand{\AWSRCU}{\mathrm{AWSRC\text{-}U}}

\newcommand{\AWSRCPF}{\mathrm{AWSRC\text{-}P}_{F}}
\DeclareMathOperator*{\argminop}{arg\,min}
\tikzset{
  block/.style={rounded corners=2pt,draw=black!55,thick,fill=blue!5,
    align=center,inner sep=3pt,minimum height=9mm,font=\small},
  ours/.style={rounded corners=2pt,draw=red!60!black,thick,fill=red!6,
    align=center,inner sep=3pt,minimum height=9mm,font=\small},
  side/.style={draw=black!35,dashed,rounded corners=2pt,fill=black!3,
    align=center,inner sep=2pt,font=\small},
  fl/.style={-{Latex[length=2mm]},thick,draw=black!60},
}

\title{Activation-Weighted Seeded Residual Coding for Low-Bit LLM Weight Repair}

\name{Zehao Liu, Chuangchuang Fang, Yang Ren}
\address{Huawei Technologies Research \& Development (UK) Limited, Edinburgh, United Kingdom\\
\texttt{zehao.liu1@h-partners.com}\\
\texttt{\{fangchuangchuang, renyang1\}@huawei.com}}

\begin{document}
\ninept
\maketitle

\begin{abstract}
Low-bit weight quantization saves storage but leaves errors that degrade LLM quality. We introduce activation-weighted seeded residual coding (AWSRC), a compact repair codec for an existing quantization backbone. Given a reconstructed weight $W_0$, AWSRC encodes the residual $W-W_0$ using deterministic seed-generated bases. The sidecar stores seed selectors, low-bit coefficients, and scales rather than an explicit codebook. Two variants combine activation weighting with per-module byte quotas ($\AWSRCU$), or blended activation/Fisher weighting with globally ranked progressive prefixes ($\AWSRCPF$) that support multiple byte budgets without refitting. On Qwen2.5-3B-Instruct, adding $0.162$ scope-bits/weight to an RTN-INT4 baseline closes $88.2\%$, $78.9\%$, and $71.3\%$ of the PPL, KL, and 11-task mean-accuracy gaps to BF16, respectively. AWSRC achieves the highest mean downstream accuracy in byte-matched residual-codec ablations and improves all metrics across model families with up to 32B parameters.
\end{abstract}


\section{Introduction}
\label{sec:intro}

Large language models (LLM) are expensive to store, transfer, and load because their weights contain billions of parameters. Weight-only post-training quantization (PTQ) reduces memory use without retraining, making it attractive for deploying existing checkpoints under fixed memory budgets. Simple round-to-nearest (RTN) quantization is inexpensive and efficient, while GPTQ and AWQ use calibration to preserve more quality \cite{gptq,awq}. These lossy quantization methods make different accuracy-cost trade-offs, but each produces a reconstructed weight $W_0$ that differs from the original full-precision $W$. The remaining error $R=W-W_0$ is useful information. Adding it back exactly would recover $W$, but a dense high-precision copy of $R$ is as costly as the weight it repairs. A practical residual code must therefore recover model quality with a small and explicitly measured payload. Errors should be measured by their impact on layer outputs, while all methods should be compared under the same end-to-end storage budget, including metadata and padding.

We study this problem as \emph{post-hoc residual repair}. The backbone quantizer and $W_0$ remain fixed, while a separately serialized sidecar approximates $R$. Therefore, repair can be attached to a cheap or already available backbone. The backbone and repaired model can be evaluated as a matched pair, so the quality gain and additional bytes are directly attributable to the residual codec.

Existing residual-recovery methods struggle to provide compact correction when the quantization residual is neither guaranteed to be sparse nor well approximated at low rank. Sparse correction brings coordinate overhead \cite{squeezellm}, low-rank correction stores dense factors \cite{lqer,qera}, and vector quantization stores assignments together with a learned codebook \cite{aqlm,gptvq}. To address this limitation, we introduce seeded coding as a compact alternative to an explicit basis or codebook. The encoder and decoder share a deterministic generator, so a short seed identifies reproducible basis atoms. SeedLM applies this principle to full weight blocks \cite{seedlm}; instead, we retain the reconstruction of an off-the-shelf quantizer and seed-code only its residual, reducing the coding target to a smaller-magnitude signal relative to a fixed base quantizer. For better compression, we weight residual errors by activation statistics, prioritizing those most relevant to layer outputs. We further require the sidecar to be independently budgeted, deterministically decodable, and backbone-compatible.

Accordingly, we propose \textbf{AWSRC}, an activation-weighted seeded residual codec. AWSRC partitions $R$ into tiles, generates candidate bases from seed-dependent sign flips and selected Hadamard columns, and fits low-bit coefficients under an activation-weighted least-squares objective.
Rather than correcting all tiles uniformly, AWSRC ranks residual corrections by activation-weighted error reduction per serialized byte. Each record stores a tile index, seed selector, low-bit coefficients, and scale. The decoder regenerates and adds the selected correction without modifying the backbone. Ordering records by gain also produces decodable prefixes at multiple rates. Since AWSRC depends only on the reconstructed backbone, the same codec supports both RTN and calibrated backbones. Our contributions are threefold.

\begin{list}{\textbullet}{%
  \setlength{\leftmargin}{1.1em}%
  \setlength{\labelwidth}{0.8em}%
  \setlength{\labelsep}{0.3em}%
  \setlength{\itemsep}{1pt}%
  \setlength{\parsep}{0pt}%
  \setlength{\topsep}{2pt}%
  \setlength{\partopsep}{0pt}}
\item A zero-codebook detachable residual codec that repairs a fixed low-bit backbone with
seed-generated bases and compact low-bit records.
\item Activation-weighted fitting and byte-normalized allocation, with optional Fisher weighting and progressive prefixes.
\item A byte-audited five-model evaluation, including an 11-task downstream study on Qwen2.5-3B. At $+0.162$ scope-bpw, AWSRC recovers $88.2\%$, $78.9\%$, and $71.3\%$ of the matched PPL, KL, and accuracy gaps to BF16.
\end{list}

\begin{figure*}[!t]
\centering
\includegraphics[width=\textwidth]{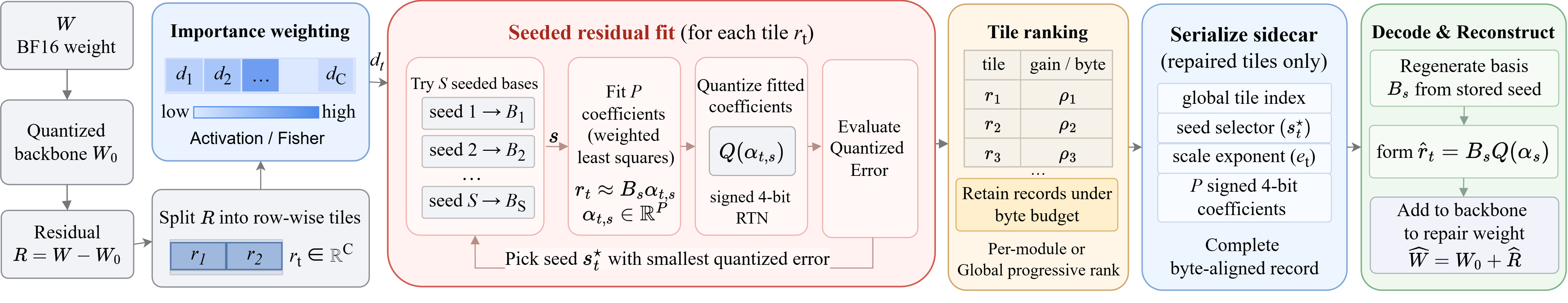}
\caption{AWSRC augments an arbitrary quantization backbone. It codes $R=W-W_0$ per tile with a seed-generated basis and an activation-weighted fit. Fisher weighting and record ordering are optional.}
\label{fig:pipeline}
\end{figure*}

\section{Related work}

Weight-only PTQ improves the primary quantized representation. GPTQ compensates sequential rounding with second-order information, while AWQ rescales activation-salient channels \cite{gptq,awq}.
The recent QAM-W method combines activation-aware scaling, Hadamard rotation, and learned two-dimensional codes \cite{qamw}. We independently reimplement QAM-W from its paper specification and use it as a strong calibrated backbone. Prior work has explored several approaches to residual recovery. Sparse formats retain sensitive outliers and their coordinates \cite{squeezellm}; LQER and QERA reconstruct quantization error with dense low-rank factors \cite{lqer,qera}; and AQLM and GPTVQ learn vectors and assignments \cite{aqlm,gptvq}. SeedLM represents complete weight blocks using bases regenerated from shared seeds, avoiding an explicit codebook \cite{seedlm}. AWSRC changes both the coding target and its role: seeded corrections are applied only to the residual of a fixed, independently chosen backbone $W_0$, with records prioritized by activation-weighted repair gain per byte. Section~\ref{sec:byte} compares these residual representations under matched backbone, repaired-scope, and serialized-sidecar budgets. Any-Precision and BitStack instead construct a single representation that supports multiple model precisions \cite{anyprecision,bitstack}. AWSRC leaves the backbone representation intact and orders complete residual records, allowing prefixes to provide fine-grained, non-integer scope-bpw without replacing the backbone codec.

\section{AWSRC Method}
\label{sec:method}

Figure~\ref{fig:pipeline} summarizes AWSRC. It computes $R=W-W_0$ from the original weight and reconstructed quantized backbone, splits $R$ into tiles, and weights their coordinates by activation importance. For each tile, multiple seeds deterministically generate candidate pseudo-random bases, each with $P$ fitted coefficients that are subsequently quantized. AWSRC retains the seed with the lowest post-quantization weighted residual error, ranks tile corrections by gain per byte, and stores the selected records under the byte budget. The sidecar contains only tile indices, seed identifiers, scale exponents, and quantized coefficients. At reconstruction, each stored seed regenerates its basis, the quantized coefficients reconstruct the residual correction, and the correction is added to $W_0$ to form the repaired weight. Fisher weighting and progressive ordering are optional.

\subsection{Seeded residual representation}

For dense target weight $W$ and a quantizer $Q$, AWSRC defines
\begin{equation}
 W_0=Q(W),\quad R=W-W_0,\quad \widehat W=W_0+\widehat R.
 \label{eq:backbone}
\end{equation}
Here $W_0$ is the reconstructed backbone, $R$ and $\widehat R$ are the target and coded residuals, and $\widehat W$ is the repaired weight. The residual is split into row-wise tiles $r_t\in\mathbb{R}^{C}$. Seed $s$ generates $B_s\in\mathbb{R}^{C\times P}$; tile- and seed-specific quantized coefficients $\widehat\alpha_{t,s}$ reconstruct the candidate correction $\widehat r_{t,s}$ as
\begin{equation}
 B_s=\operatorname{GenBasis}(s,C,P),\qquad
 \widehat r_{t,s}=B_s\widehat\alpha_{t,s} .
 \label{eq:basis}
\end{equation}
We use normalized signed and permuted Hadamard bases. A seed selects a sign pattern, coordinate permutation, and $P$ columns of the order-$C$ Hadamard matrix. Encoder and decoder share the generator and numeric convention, so only the selector is stored. Here $C$ controls locality, $P$ the subspace dimension, $b$ the coefficient precision, and $S$ the basis diversity and search cost.

\subsection{Activation-weighted fitting and seed selection}

Calibration estimates activation weights $a_k=\mathbb{E}[x_k^2]$ and $D=\operatorname{diag}(a)$. For tile $t$, let $x_{t,j}$ and $d_{t,j}$ denote its aligned input coordinate and weight. For every candidate seed $s$, we fit
\begin{equation}
 \alpha_{t,s}^\star=\argminop_{\alpha}
 \sum_{j=1}^{C}d_{t,j}\bigl([B_s\alpha]_j-r_{t,j}\bigr)^2
 +\epsilon\|\alpha\|_2^2 .
 \label{eq:weighted-ls}
\end{equation}
Here $\alpha_{t,s}^\star$ is the floating-point fit and $\epsilon>0$ is a ridge constant. For every $(t,s)$, we quantize $\alpha_{t,s}^\star$ by coefficient-wise 4-bit RTN with one power-of-two scale per tile and reconstruct $\widehat r_{t,s}$. The encoder chooses $s_t^\star$ by the lowest post-quantization weighted error and sets $\widehat r_t=\widehat r_{t,s_t^\star}$. Thus selection reflects the stored coefficients rather than the unquantized least-squares fit. Under the diagonal activation-covariance approximation, the weighted sum estimates the expected output perturbation, $\mathbb{E}_{x_t}[((r_t-\widehat r_{t,s})^\top x_t)^2]\approx\sum_{j=1}^{C}d_{t,j}(r_{t,j}-\widehat r_{t,s,j})^2$. Activations determine important residual errors but are not stored. The activation diagonal is shared across output rows; the optional Fisher blend supplies row-specific weights. Each record stores a tile index, seed selector, $P$ signed $b$-bit coefficients, and a power-of-two scale exponent. Decoding regenerates $B_{s_t^\star}\widehat{\alpha}_{t,s_t^\star}$ and adds it to $W_0$; unselected tiles remain unchanged. Calibration statistics and seed search are encoder-only, and an empty sidecar returns $W_0$ exactly.

\subsection{Byte-aware allocation and progressive prefixes}

The encoder evaluates the reconstruction after coefficient quantization and assigns tile $t$ the byte-normalized gain
\begin{equation}
 \rho_t=\frac{1}{L_t}\sum_{j=1}^{C}d_{t,j}
 \left[r_{t,j}^2-(r_{t,j}-\widehat r_{t,j})^2\right],
 \label{eq:gain}
\end{equation}
where $L_t$ is the complete serialized record size. Non-positive records are discarded. Uniform allocation retains records within each module, whereas the progressive variant sorts all eligible records by $\rho_t$. Every prefix ending at a complete-record boundary decodes against the same $W_0$ without refitting coefficients or changing the decoder. In $\AWSRCPF$, F blends Fisher and activation diagonals, while P denotes global ranking by $\rho_t$. Separately, scale double quantization (SDQ), derived from QLoRA's Double Quantization, further compresses RTN backbone group-scale metadata \cite{qlora}. For RTN with group size 128, SDQ reduces scale metadata from 0.125 to 0.068 scope-bpw, a 45\% reduction. Under a fixed total scope-bpw, the saved metadata budget can be reassigned to AWSRC residual records.

\subsection{Serialized rate}

For payload length $\ell_{\rm payload}=\lceil\log_2 S\rceil+b_e+Pb$, where $b_e$ is the scale-exponent width, the measured storage is
\begingroup
\small
\setlength{\abovedisplayskip}{2pt}
\setlength{\belowdisplayskip}{2pt}
\begin{equation}
\begin{aligned}
B_{\rm side}&=h+K\left\lceil(\lceil\log_2T\rceil+\ell_{\rm payload})/8\right\rceil,\\[-1pt]
\mathrm{bpw}_{\rm scope}&=8(B_{\rm bb}+B_{\rm side})/N_{\rm scope}.
\end{aligned}
\label{eq:storage}
\end{equation}
\endgroup
Here $T$ and $K$ are candidate and retained tile counts, $h$ is the stream-header size, $B_{\rm bb}$ is the serialized backbone size, and $N_{\rm scope}$ is the number of parameters in the repair scope. Scope-bpw measures storage over the repaired matrices. Full-effective-bpw is the serialized model-wide weight payload in bits divided by the total parameter count; for official GPTQ/AWQ checkpoints, it therefore reflects their native serialization and quantization scopes.

\begin{figure*}[!b]
\centering
\vspace{-5pt}
\includegraphics[width=\textwidth,trim=0 6pt 0 0,clip]{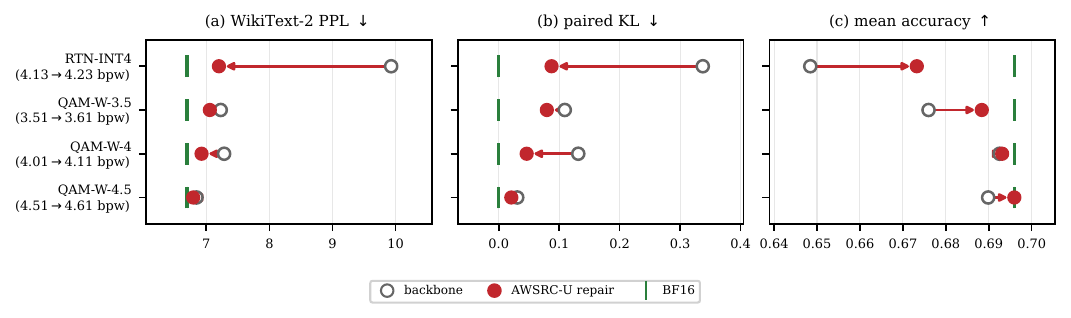}
\vspace{-5pt}
\caption{Multi-seed $\AWSRCU$ repairs at $\Delta\mathrm{bpw}_{\rm scope}=+0.10$ on Qwen2.5-3B MLP projections. QAM-W is our clean-room reproduction. Arrows point from each backbone to its repair, and green lines mark BF16.}
\label{fig:paired-breadth}
\end{figure*}

\section{Experiments}
\label{sec:experiments}

We evaluate Qwen2.5-3B-Instruct \cite{qwen25}. All experiments were conducted on a single Ascend 910C NPU with 128 GiB HBM. Both backbone quantization and AWSRC repair cover its 108 MLP projections, comprising \texttt{gate\_proj}, \texttt{up\_proj}, and \texttt{down\_proj} in every layer, with all other weights kept in BF16. Activation statistics use 256 captured input rows per projection from 32 deterministic, template-generated technical sentences. The optional Fisher diagonal uses next-token gradients from 4,096 WikiText-2 training tokens \cite{wikitext}. PPL uses 16,384 WikiText-2 test tokens, and paired KL uses 4,096 tokens with the BF16 teacher. Six-task accuracy averages PIQA \cite{piqa}, HellaSwag \cite{hellaswag}, COPA \cite{copa}, RTE \cite{glue}, OpenBookQA \cite{openbookqa}, and LAMBADA \cite{lambada}, capped at 300 examples per task. A full-split extension adds MMLU \cite{mmlu}, ARC-Easy/Challenge \cite{arc}, WinoGrande \cite{winogrande}, and BoolQ \cite{boolq}, totaling 11 tasks and 40,037 examples; all evaluations use lm-eval \cite{lmeval}.

Paired KL measures token-averaged output-distribution drift from the BF16 teacher under identical prefixes. Let $m_0$, $m_r$, and $m_b$ denote the backbone, repaired, and BF16 scores. Gap recovery is $(m_0-m_r)/(m_0-m_b)$ for lower-is-better PPL/KL and $(m_r-m_0)/(m_b-m_0)$ for accuracy. Percentages use unrounded scores from the stated matched evaluation. Accuracy is computed per task before taking the unweighted mean.
RTN uses group size 128, and optional SDQ is applied only to RTN backbones. GPTQ-W4 \cite{gptq} and AWQ-W4 \cite{awq} are official all-linear references. QAM-W \cite{qamw} is our paper-specification clean-room implementation, so we report only matched before--after comparisons. All candidates are reconstructed as dense BF16 weights for evaluation. The default calibration seed is 0, and multi-seed results use seeds 0, 1, and 2. Basis generation, tile order, calibration order, and tie breaking are fixed. Serialized rates include indices, payloads, headers, and padding. We focus on operating points around 4 scope-bpw, where our sweep gives the strongest quality--storage trade-off. For Qwen2.5, each record uses 7 bytes, the header uses 904 bytes, and the 4.23-bpw sidecar is 49.25 MB. At each rate, we globally rank positive-gain records by $\rho_t$ and retain the longest prefix within the byte budget; 4.23 scope-bpw retains $K=7{,}034{,}996$ tiles. The complete 2.60 GB Qwen2.5-3B checkpoint can be independently loaded with identical reconstructed weights and less than 0.01 PPL difference from in-memory reconstruction.

$\AWSRCU$ uses activation weights, per-module quotas, and $(C,P,b,S)=(128,2,4,2)$. $\AWSRCPF$ uses global progressive ranking and $(C,P,b,S)=(16,4,4,16)$. Parameter sweeps select $\AWSRCU$ as the simplest low-cost configuration and $\AWSRCPF$ as the best-quality configuration. For $\AWSRCPF$, a blend sweep in increments of $0.25$ selects $0.75$ Fisher and $0.25$ activation weighting. These are distinct cost--quality configurations rather than a controlled one-factor ablation.

\subsection{Main results}

Figure~\ref{fig:paired-breadth} compares $\AWSRCU$ at a fixed $+0.1$ scope-bpw on four matched backbones. Mean PPL, KL, and six-task accuracy improve for RTN-INT4 and all three QAM-W rates. The largest recovery occurs on RTN, while the smaller but consistent QAM-W changes show that the same low-cost codec can also repair stronger backbones.

Table~\ref{tab:main} separates quality references from controlled matched repairs. At the same $+0.1$ scope-bpw, $\AWSRCPF$ gives the lowest PPL and KL on both RTN backbones and all three QAM-W backbones. The two variants remain close on QAM-W-4.5, where $\AWSRCU$ gives slightly higher accuracy. Accuracy changes are smaller and vary across backbone and AWSRC configurations.

\begin{table}[H]
\centering
\caption{Qwen2.5-3B. Repair rows are multi-seed means at $+0.1$ scope-bpw. $s/f$ denotes scope/full-effective-bpw, $\dagger$ official all-linear references, and $\ddagger$ our clean-room QAM-W.}
\label{tab:main}
\small
\setlength{\tabcolsep}{2.2pt}
\renewcommand{\arraystretch}{1.00}
\begin{tabular*}{\columnwidth}{@{\extracolsep{\fill}}lrrrr@{}}
\toprule
Weight configuration & bpw $s/f$ & PPL & KL & 6-task acc. \\
\midrule
BF16 & 16.00 & 6.704 & 0.000 & 0.696 \\
GPTQ-W4$\dagger$ & 4.17/5.36 & 7.115 & 0.096 & 0.687 \\
AWQ-W4$\dagger$ & 4.16/6.97 & 7.149 & 0.089 & 0.693 \\
\midrule
RTN-INT4 & 4.13/6.63 & 9.929 & 0.338 & 0.648 \\
\quad$+\AWSRCU$ & \multirow{2}{*}{4.23/6.71} & 7.200 & 0.087 & 0.673 \\
\quad$+\AWSRCPF$ & & \textbf{7.142} & \textbf{0.076} & \textbf{0.680} \\
RTN-INT4-SDQ & 4.07/6.59 & 9.834 & 0.339 & 0.652 \\
\quad$+\AWSRCU$ & \multirow{2}{*}{4.17/6.67} & 7.180 & 0.089 & 0.677 \\
\quad$+\AWSRCPF$ & & \textbf{7.130} & \textbf{0.076} & \textbf{0.684} \\
QAM-W-3.5$\ddagger$ & 3.51/6.15 & 7.229 & 0.109 & 0.676 \\
\quad$+\AWSRCU$ & \multirow{2}{*}{3.61/6.22} & 7.056 & 0.080 & 0.688 \\
\quad$+\AWSRCPF$ & & \textbf{7.033} & \textbf{0.074} & \textbf{0.692} \\
QAM-W-4$\ddagger$ & 4.01/6.54 & 7.283 & 0.132 & 0.693 \\
\quad$+\AWSRCU$ & \multirow{2}{*}{4.11/6.62} & 6.925 & 0.046 & 0.693 \\
\quad$+\AWSRCPF$ & & \textbf{6.904} & \textbf{0.045} & \textbf{0.694} \\
QAM-W-4.5$\ddagger$ & 4.51/6.93 & 6.845 & 0.031 & 0.690 \\
\quad$+\AWSRCU$ & \multirow{2}{*}{4.61/7.01} & 6.792 & 0.021 & \textbf{0.696} \\
\quad$+\AWSRCPF$ & & \textbf{6.791} & \textbf{0.020} & 0.695 \\
\bottomrule
\end{tabular*}
\end{table}

\subsection{Byte-matched residual codecs}
\label{sec:byte}

We construct sparse \cite{squeezellm}, low-rank \cite{lqer,qera}, and learned-VQ \cite{aqlm,gptvq} residual-codec alternatives inspired by prior work, adapting each to encode $R=W-W_0$ under the same weighted objective and byte budget rather than reproducing the cited systems. All four codecs repair the same RTN-INT4-SDQ backbone and 108 matrices. Every sidecar is exactly 49,245,876 bytes including headers and padding, giving 4.23 scope-bpw and 6.71 full-effective-bpw. Sparse coding stores FP16 values and 32-bit coordinates, low-rank coding uses activation-weighted rank-34 INT8 factors, and VQ uses per-module $K=256$ FP16 codebooks with uint8 assignments.

\begin{table}[H]
\centering
\caption{Byte-matched residual-codec alternatives on an RTN-INT4-SDQ backbone. Values are mean $\pm$ sample s.d. over 10 calibration seeds (0--9). Bold marks the best unrounded mean.}
\label{tab:bytes}
\small
\setlength{\tabcolsep}{0.7pt}
\renewcommand{\arraystretch}{1.00}
\begin{tabular}{@{}lrrr@{}}
\toprule
Codec & PPL & KL & 6-task acc. \\
\midrule
Sparse & $7.145\pm0.003$ & $0.078\pm0.001$ & $0.682\pm0.002$ \\
Quantized low-rank & $7.083\pm0.006$ & $0.073\pm0.001$ & $0.675\pm0.003$ \\
Learned VQ & $\mathbf{7.072}\pm0.014$ & $\mathbf{0.068}\pm0.001$ & $0.675\pm0.002$ \\
Seeded & $7.077\pm0.003$ & $0.072\pm0.000$ & $\mathbf{0.692}\pm0.002$ \\
\bottomrule
\end{tabular}
\end{table}

With backbone, scope, and bytes fixed, Table~\ref{tab:bytes} isolates codec efficiency. VQ gives slightly lower mean PPL and KL but has the largest variation in both. We evaluate seeded coding with $\AWSRCPF$, obtaining the lowest PPL/KL variability and highest mean accuracy. These results support AWSRC's seeded design without an explicit codebook.

\subsection{Downstream task and model generalization}
\label{sec:generalization}
\begin{figure}[H]
\centering
\includegraphics[width=\columnwidth,trim=0 9pt 0 7pt,clip]{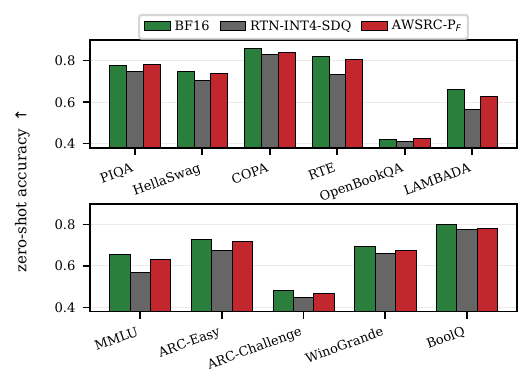}
\vspace{-5pt}
\caption{Full-split zero-shot accuracy on 11 Qwen2.5-3B tasks.}
\label{fig:per-task}
\vspace{-5pt}
\end{figure}

On the 11 full task splits, $\AWSRCPF$ at 4.23 scope-bpw changes PPL/KL from $9.834/0.339$ to $7.075/0.071$ and accuracy from $0.648$ to $0.682$, closing $88.2\%$, $78.9\%$, and $71.3\%$ of the BF16 gaps. Figure~\ref{fig:per-task} shows consistent task-level movement toward BF16 rather than recovery driven by only a few tasks.

\vspace{-8pt}

\begin{table}[H]
\centering
\caption{Cross-model repair at 4.23 scope-bpw. PPL, KL, and acc. are measured after adding $\AWSRCPF$ to RTN-INT4-SDQ; each $\Delta$ is the repaired value minus the uncorrected backbone value.}
\label{tab:transfer}
\small
\setlength{\tabcolsep}{1.0pt}
\renewcommand{\arraystretch}{1.00}
\begin{tabular*}{\columnwidth}{@{\extracolsep{\fill}}lrrrrrr@{}}
\toprule
Model & PPL & $\Delta$PPL & KL & $\Delta$KL & acc. & $\Delta$acc. \\
\midrule
Qwen3-32B \cite{qwen3} & 6.073 & -4.776 & 0.105 & -0.660 & 0.723 & +0.045 \\
Qwen3-4B \cite{qwen3} & 8.101 & -0.208 & 0.048 & -0.040 & 0.676 & +0.016 \\
Qwen2.5-14B \cite{qwen25} & 3.814 & -0.132 & 0.310 & -0.061 & 0.754 & +0.008 \\
Llama-3.2-3B \cite{llama32} & 9.807 & -0.547 & 0.044 & -0.031 & 0.668 & +0.001 \\
Yi-1.5-9B \cite{yi} & 5.244 & -0.024 & 0.064 & -0.016 & 0.704 & +0.006 \\
Mistral-7B \cite{mistral7b} & 4.722 & -0.013 & 0.028 & -0.003 & 0.725 & +0.002 \\
\bottomrule
\end{tabular*}
\end{table}

Table~\ref{tab:transfer} lists metrics of the repaired RTN-INT4-SDQ backbones, with each $\Delta$ defined as repaired minus backbone; acc. denotes six-task mean accuracy. Under the fixed $\AWSRCPF$ setting, PPL, KL, and mean accuracy improve across all six models. Qwen3-32B shows the largest PPL and KL reductions, recovering $97.1\%$ of the PPL gap and reducing KL to the BF16 teacher by $86.2\%$. The smaller but consistent gains on Mistral-7B indicate that repair headroom varies with the backbone while remaining positive under a shared setting.

\section{Conclusion}

AWSRC repairs quantized weights using seeded bases and gain-per-byte allocation, without storing an explicit codebook or dense correction. $\AWSRCU$ provides activation-weighted per-module allocation; $\AWSRCPF$ combines activation/Fisher weighting with global ranking and nested prefixes that support multiple budgets without refitting. These results demonstrate that compact seeded residuals improve the storage--quality trade-off of existing quantization backbones without redesigning the backbone or storing a dense correction. AWSRC achieves the highest mean downstream accuracy among byte-matched residual codecs. A fixed setting improves PPL, KL, and mean accuracy across six models up to 32B parameters. Future work will evaluate more models and parameter combinations and develop packed implementations for practical deployment.

\clearpage
\bibliographystyle{IEEEbib}
\bibliography{references}

\section{Compliance with Ethical Standards}
This computational study used publicly available model checkpoints and benchmark datasets and involved no human or animal subjects. No ethical approval was required. The authors declare no conflicts of interest.
\end{document}